# Shared Autonomous Vehicle Mobility for a Transportation Underserved City

**Karina Meneses-Cime, Bilin Aksun-Guvenc, Levent Guvenc**
Automated Driving Lab, Ohio State University

## Abstract

This paper proposes the use of an on-demand, ride hailed and ride-Shared Autonomous Vehicle (SAV) service as a feasible solution to serve the mobility needs of a small city where fixed route, circulator type public transportation may be too expensive to operate. The presented work builds upon our earlier work that modeled the city of Marysville, Ohio as an example of such a city, with realistic traffic behavior, and trip requests. A simple SAV dispatcher is implemented to model the behavior of the proposed on-demand mobility service. The goal of the service is to optimally distribute SAVs along the network to allocate passengers and shared rides. The pickup and drop-off locations are strategically placed along the network to provide mobility from affordable housing, which are also transit deserts, to locations corresponding to jobs and other opportunities. The study is carried out by varying the behaviors of the SAV driving system from cautious to aggressive along with the size of the SAV fleet and analyzing their corresponding performance. It is found that the size of the network and behavior of AV driving system behavior results in an optimal number of SAVs after which increasing the number of SAVs does not improve overall mobility. For the Marysville network, which is a 9 mile by 8 mile network, this happens at the mark of a fleet of 8 deployed SAVs. The results show that the introduction of the proposed SAV service with a simple optimal shared scheme can provide access to services and jobs to hundreds of people in a small sized city.

## Introduction

Research on active safety systems and ADAS [1], [2], [3], [4], [5] has evolved naturally to connected vehicle (CV) and autonomous vehicle (AV) research including robust and fuel optimal controls [6], [7] and automated driving algorithms [8], [9], [10], [11], [12] resulting in the successful deployment of AV shuttles and taxis [13], [14]. The introduction of Autonomous Vehicle (AV) technologies has been shown to bring about positive side effects to network capacity and traffic flow, and is associated with improved fuel/energy economy [15], [16], [17], [18], [19]. While being very promising, the use of autonomous vehicles alone will not be a solution to improve mobility and traffic flow as long as there are only one or two occupants in the vehicle. On-demand, ride hailed and ride-Shared Autonomous Vehicles (SAVs) are, therefore, the focus of this paper as they offer an economically feasible, hence sustainable, mobility solution. These SAVs should have a larger occupancy count of five or more passengers. For optimal operation in picking up passengers from different locations, their routing should be optimized to increase the number of passengers served and vehicle occupancy while decreasing trip time. Well known minimum distance path optimization algorithms such as the A* and Dijkstra algorithms will not be enough to handle this required optimization [20] which should be able to balance the number of passengers served versus trip times.

Reference [21] uses a grid based approach to show how the introduction of SAV technologies can benefit the traffic network in the form of positive emission impacts at the cost of travel distance per SAV. In reference [22], through the modeling of Lisbon, Portugal based on a grid network, researchers found that replacing public transport with Autonomous Taxis (ATs) meant that only 3% of the size of public transportation was needed to cover transportation demand needs. It was also seen that traffic congestion and undesired emissions decreased and there was more available space for parking. Further, reference [23] studied the introduction of SAVs at a low market penetration level of regional trips in a network based on Austin, Texas. It was found that an SAV could replace nine conventional vehicles and 8% more vehicle miles traveled (VMT) could be generated. Reference [24] models a small network in Budapest via grid-simulation and network link simulation using SUMO and shows capacity improvement proportional to AV penetration rate. The studies cited above provide insight for a future with AVs by generating data for validation. However, though the grid-based approach addresses the large computational demands necessary for large simulations, the simulation fidelity of each agent is reduced, showing the need for analysis of single-agent performance in planning and simulating an SAV deployment. In contrast to the existing literature, this paper utilizes a microscopic traffic simulation, building upon our earlier work in [25] and extending it to optimize an SAV service in a small city. This SAV service is on-demand and ride-hailed making it an SAV mobility on demand solution.

While non-AV mobility on-demand services are common today, future deployments will replace the operator with an an AV driver which will lead to decisions on how to dispatch according to incoming requests. While driving a regular taxi, drivers may unknowingly optimize the trips based on their earnings. However, with AV technologies, there is a chance of optimizing based on AV cooperation. There have been many advances towards driverless operation of autonomous vehicles used as taxis in the U.S. and around the world with many successful deployments in current operation. The future elimination of the driver based on this trend of operating driverless robo-taxis means that the vehicles can be operated for longer periods of time in between charging or fueling stops. As a result, the number of privately owned vehicles is expected to decrease along with an increase in the number of AVs operating on the road.

Driverless operation of on-demand and ride hailed SAVs has the potential ro improve mobility choices of especially the diverse, low-income and underserved communities including the elderly and disabled people. This paper focuses on such a solution and formulates and presents optimization for improving its operation to maximize the number of shared miles and provide lower-resource communities an affordable mobility choice for access to jobs and other locations of opportunity.

The outline of the rest of the paper is as follows. The modeling of the traffic network of the city of Marysville in Ohio and the method of microscopic traffic simulation are reviewed briefly next. The paper, then, presents a simulation study on optimally distributing SAVs along the network to allocate passengers and shared rides. The study is carried out by varying the behaviors of the SAV driving system from cautious to aggressive along with the size of the SAV fleet and analyzing their corresponding performance. The paper ends with conclusions.

## Marysville Network

The City of Marysville, Ohio was chosen as the city under consideration here as it does not have public transportation. This is due to the fact that a fixed route and fixed schedule circulator type bus service is too expensive to operate. So, an on-demand ride-hailed SAV mobility service is a promising and economically feasible solution. To achieve the goal of providing mobility to the residents of Marysville living in affordable housing in the periphery of the city, the network was divided into several parts first as shown in Figure 1 where the colored area at the center shows the location of jobs, schools, shopping, medical facilities and other places of opportunity.

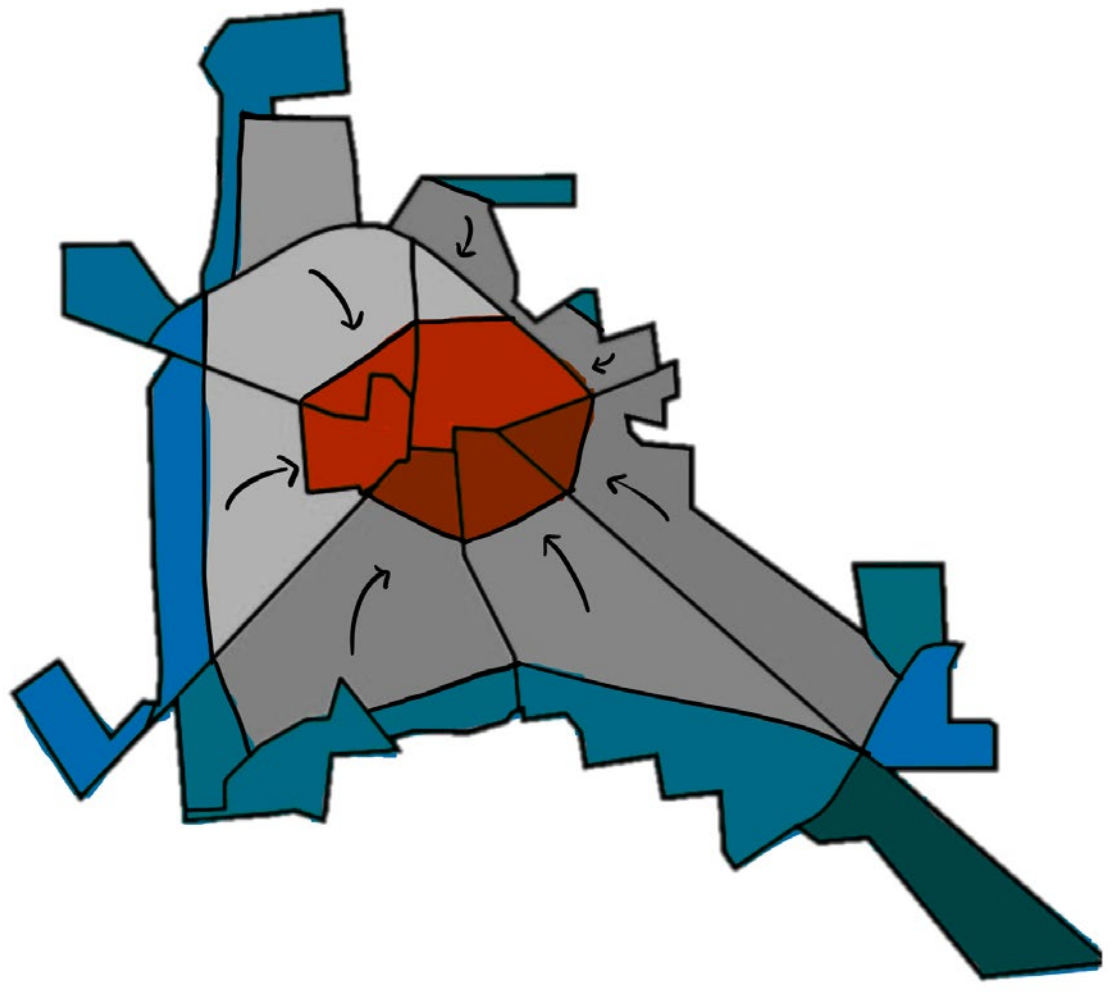

*Figure 1: SAV stop targets [25].*

The network was modeled using the microscopic traffic simulator Vissim. The network was first created using the available map in OpenStreetMap (OSM). The network was fine tuned using the approach in our earlier work using puplicly available data from the Mid Ohio Regional Planning Commission and the Ohio Department of Transportation [25]. Three different levels of autonomous driving were used as cautious, normal and aggressive [25]. An SAV demand manager was used to maximize the number of shared miles per passenger. The mathematical representation of SAV dispatching is presented next.

**Definition 1. (Directed Graph)** A directed graph $\boldsymbol{G}$ is an ordered pair of vertices $\boldsymbol{V} = \{\boldsymbol{v_i}\}$ and edges $\boldsymbol{E} = \{\boldsymbol{e_j}\}$**.**

The Marysville network is represented as a directed graph with weights on the edges. The graph is denoted by $\boldsymbol{G} = (\boldsymbol{V}, \boldsymbol{E})$. Note that each edge is simply a tuple of vertices $\boldsymbol{e_j} = (\boldsymbol{v_{j_1}}, \boldsymbol{v_{j_2}})$ that indicate a path from vertex $\boldsymbol{v_{j_1}}$ to vertex $\boldsymbol{v_{j_2}}$. In this way, streets are represented as edges and intersections as nodes. Further, each vertex is simply a tuple that represents a point in 2D space, i.e. $\boldsymbol{v_j} = (\boldsymbol{x}, \boldsymbol{y})$. The weight on the edges of $\boldsymbol{G}$ is then the Euclidean distance between the two nodes. That is, edge $\boldsymbol{e_j} = (\boldsymbol{v_{j_1}}, \boldsymbol{v_{j_2}}) = ((\boldsymbol{x_1}, \boldsymbol{y_1}), (\boldsymbol{x_2}, \boldsymbol{y_2}))$,

$$\boldsymbol{w}(\boldsymbol{e_j}) = \sqrt{(\boldsymbol{x_2} - \boldsymbol{x_1})^2 + (\boldsymbol{y_2} - \boldsymbol{y_1})^2}. \qquad (4)$$

Furthermore, the graph is assumed to be connected. Throughout the network, $\boldsymbol{m}$ pickup and dropoff locations are placed along the edges of $\boldsymbol{G}$, $\boldsymbol{B} = \{\boldsymbol{b_1}, \dots, \boldsymbol{b_m}\}$. Consider, a pickup/dropoff stop $\boldsymbol{b_i} \in \boldsymbol{B}$, then the placement of $\boldsymbol{b_i}$ along edge $\boldsymbol{e_j}$ requires the definition of distance $\boldsymbol{d}$ such that $\boldsymbol{b_i}$ is placed $\boldsymbol{d}$ meters away from the source node $\boldsymbol{v_{j_1}}$ **as shown in Figure 2.**

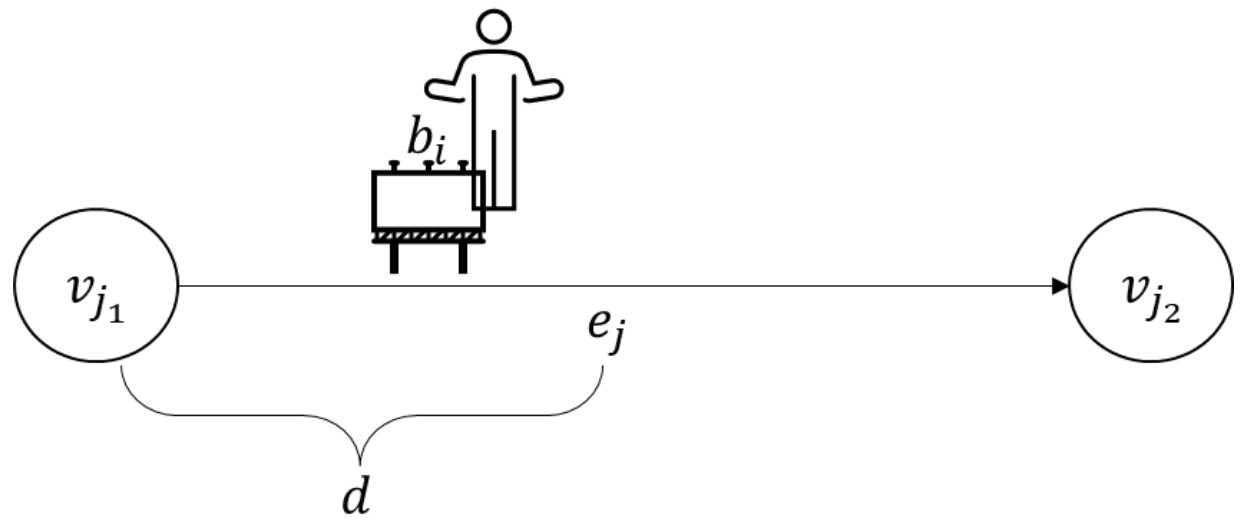

*Figure2: Visual representation of the computation of the slack distance where pickup/dropoff location where $b_i$ is placed along edge $e_j$ [25].*

The notion of distance between pickup/dropoff locations is now established. Assume that pickup/dropoff location $\boldsymbol{b_i}$ is placed along edge $\boldsymbol{e_i}$ and pickup/dropoff location $\boldsymbol{b_j}$ is placed along edge $\boldsymbol{e_j}$. At the beginning of the simulation, the list of pickup/dropoff locations creates another list of source nodes $(\boldsymbol{v_1}, \dots, \boldsymbol{v_m})$. Then, Dijkstra's algorithm [23] is run to create a list of the shortest paths between each node. Since there are $\boldsymbol{m}$ nodes, and order matters, that is, the shortest path from $\boldsymbol{v_i}$ to $\boldsymbol{v_j}$ is not necessarily the shortest path from $\boldsymbol{v_j}$ to $\boldsymbol{v_i}$, this creates a list of length $\boldsymbol{m}(\boldsymbol{m} - \boldsymbol{1})$. Further, because the graph is connected, there will be at least one shortest path between each set of nodes. When there are two shortest paths, one is picked at random (this could be done based on average traffic speed or fuel/energy economy comparisons in the future). Thus, there exists a list of shortest paths $(\boldsymbol{W_1}, \dots, \boldsymbol{W_{m(m-1)}})$ between the nodes. Denote each path by its set of edges, $\boldsymbol{W_i} = (\boldsymbol{e_{i_1}}, \dots, \boldsymbol{e_{i_k}})$. Then, for pickup/dropoff bases $\boldsymbol{b_i}$ and $\boldsymbol{b_j}$ located at distances $\boldsymbol{d_i}$ and $\boldsymbol{d_j}$ away from the source node, and with shortest path $\boldsymbol{W_i}$, by means of traveling from $\boldsymbol{b_i}$ to $\boldsymbol{b_j}$, the distance is defined as

$$\boldsymbol{d}(\boldsymbol{b_i}, \boldsymbol{b_j}) = \sum_{\boldsymbol{r=2}}^{\boldsymbol{k}} \boldsymbol{w}(\boldsymbol{e_{i_r}}) + (\boldsymbol{w}(\boldsymbol{e_{i_1}}) - \boldsymbol{d_i}) + \boldsymbol{d_j} \qquad (5)$$

The distance between pickup/dropoff locations is utilized to distpatch incoming requests. The SAVs are dispatched to passengers in order of incoming requests and distance. First, the current requests assigned to each SAV are taken into account. If there are current requests, the dispatcher checks if the passengers have waited for more than a set threshold. This threshold is set to 20 min in the simulations. The dispatcher also first prioritizes requests that have been placed but not attended to using this threshold, within a set radius. If none of these conditions are met, the dispatcher moves on to attending requests as first-come, first-serve. This logic maximizes the total number of shared miles while maintaining a handle on the quality of service by maintaining the soft deadline requirements. This sharing behavior allows for SAVs to be assigned requests at the time of trip of passengers and increasing the capacity of such.

## Simulation Study

The SAVs were simulated inside the created Marysville network for a period of 20 simulation runs per simulation scenario. For the density simulation plots, the key is provided in Figure 3. First, a baseline was obtained to observe the behavior of the network without any injected SAVs. Throughout the simulation, the density of the network remained mostly equal. This can be visualized in Figure 4.

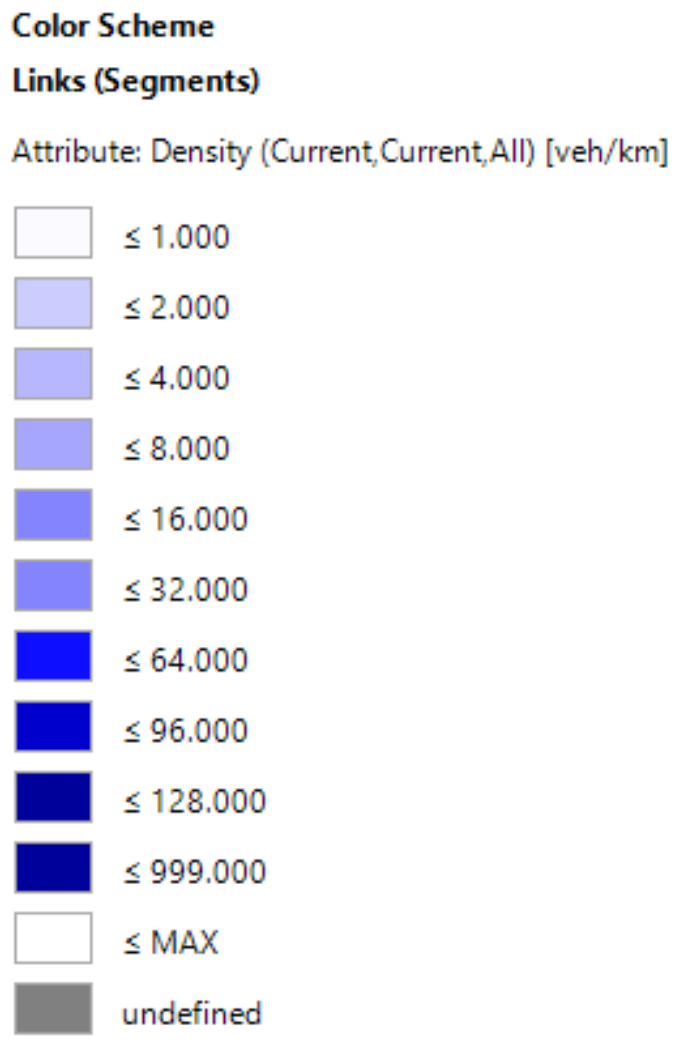

*Figure 3: Density simulations key.*

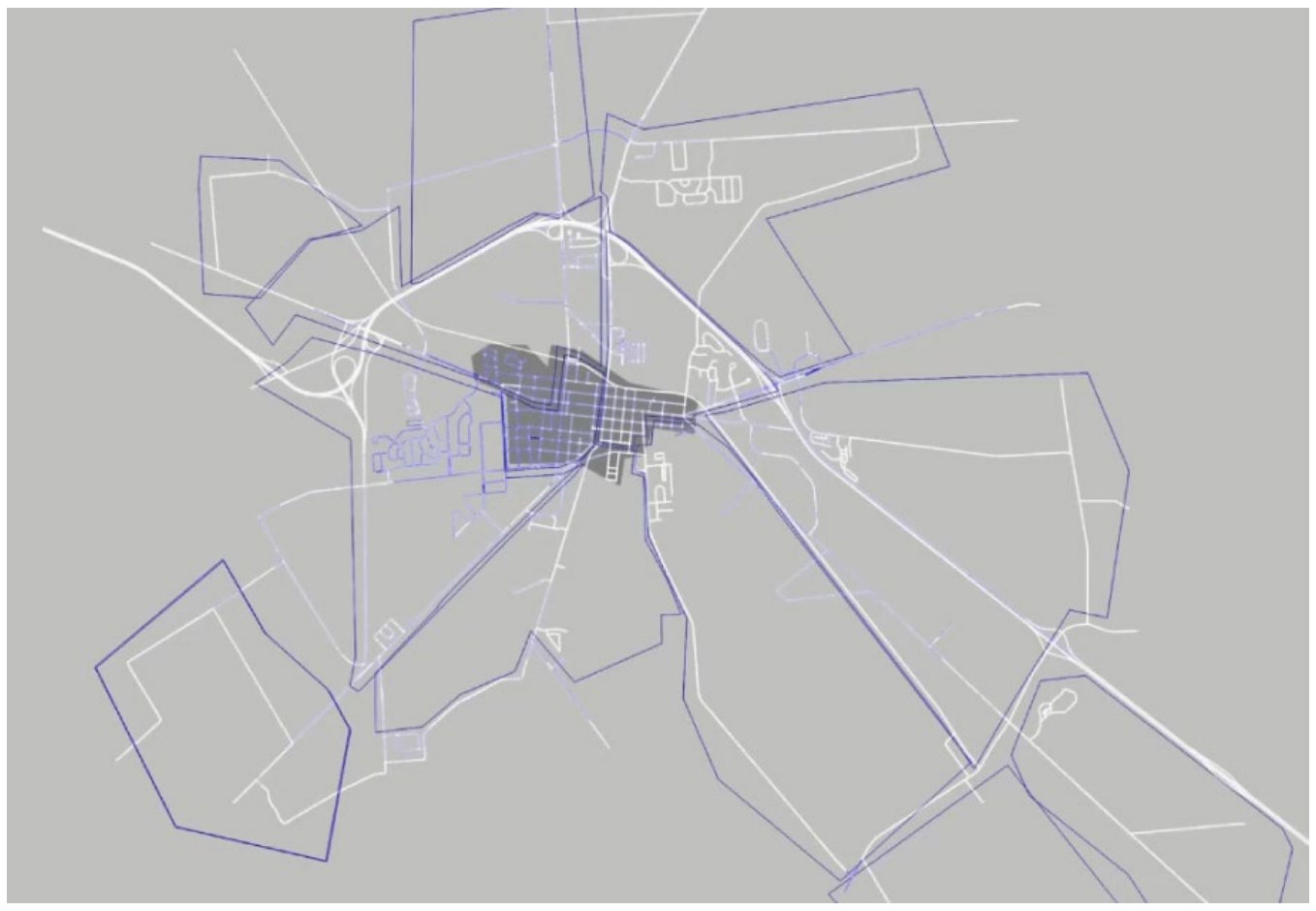

*Figure4: Marysville Network without SAVs.*

Next, 10 SAVs were injected into the network such that they provided service to the peripheral housing locations. The simulation during the first phase of deploying the SAVs can be seen in Figure 5. Then, the SAVs provide movement to the peripheral areas of the network, resulting in Figure 6.

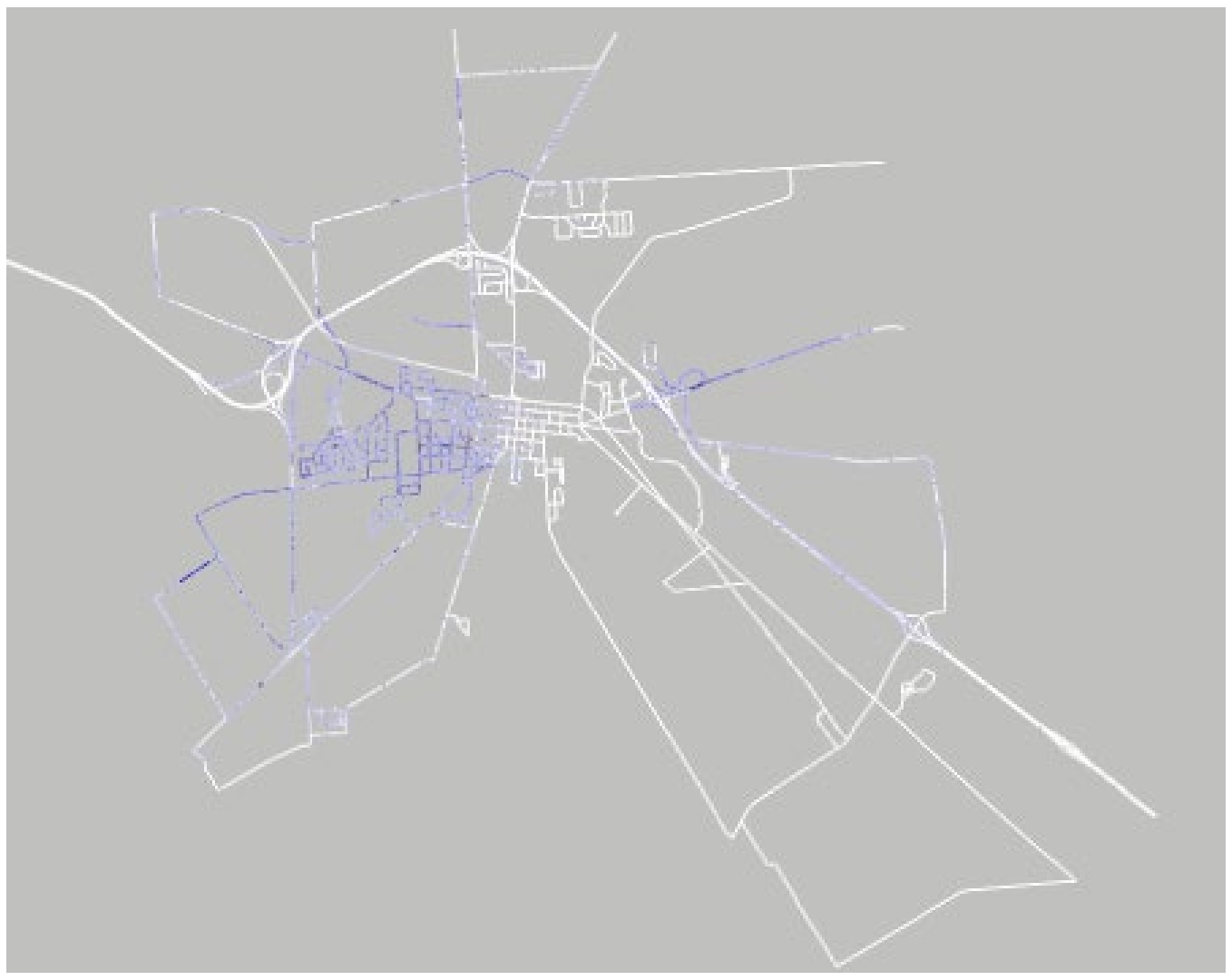

*Figure 5: Beginning phase of SAV deployment.*

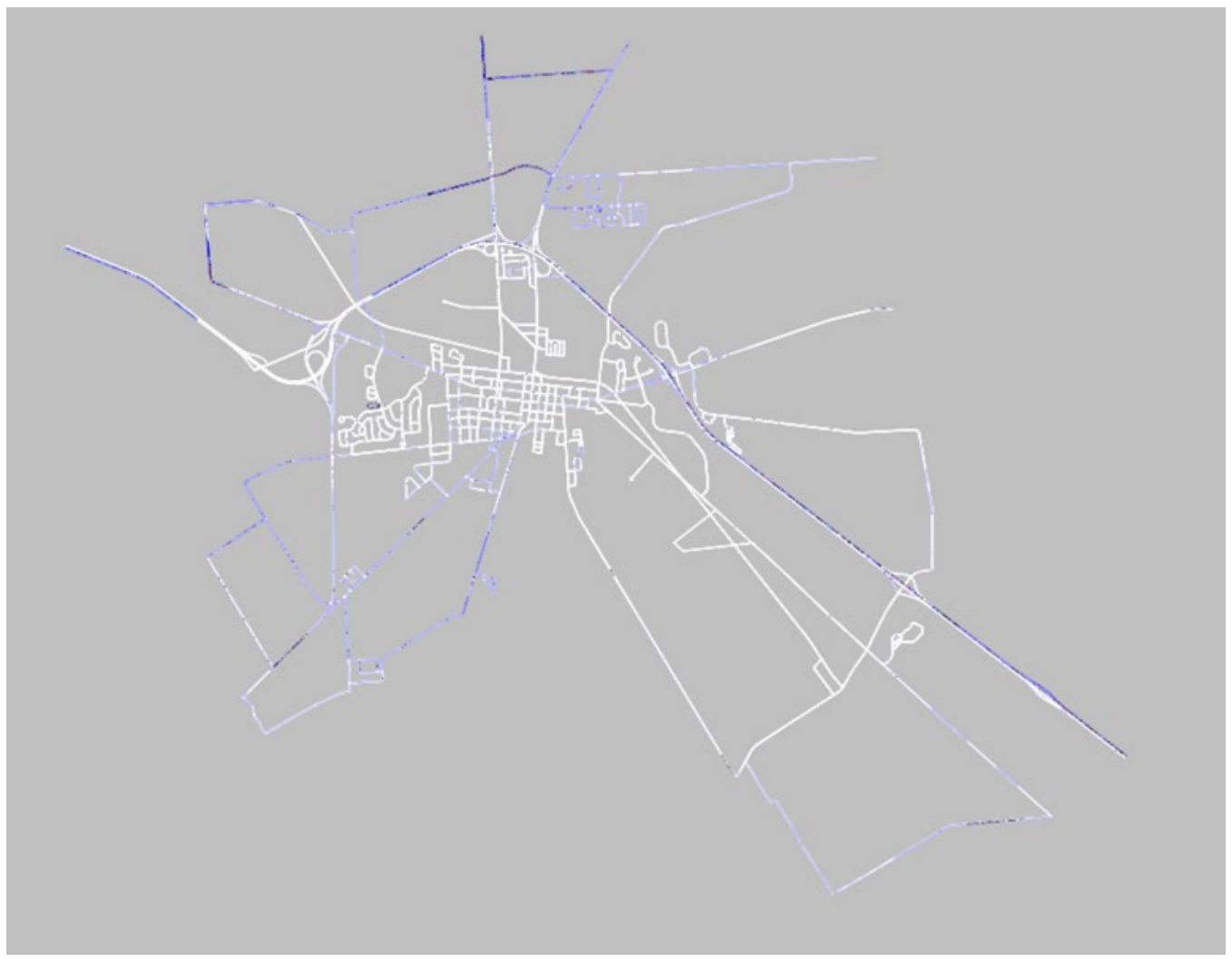

*Figure 6: Second phase of SAV deployment. SAV drop-offs.*

The traffic network is assessed further in the following. It was found that with only 2 injected SAVs, the average delay per vehicle increased for 20 minutes but the average number of stops per vehicle decreased to almost non-existent. This can be attributed to a smoothing out in the traffic flow of the network due to the deterministic behavior of the AV drivers. The average delay per vehicle can be seen in Figure 7 and the average number of stops per vehicle can be seen in Figure 8. However, the overall distance traveled increased by the addition of the SAVs as observed in Figure 9.

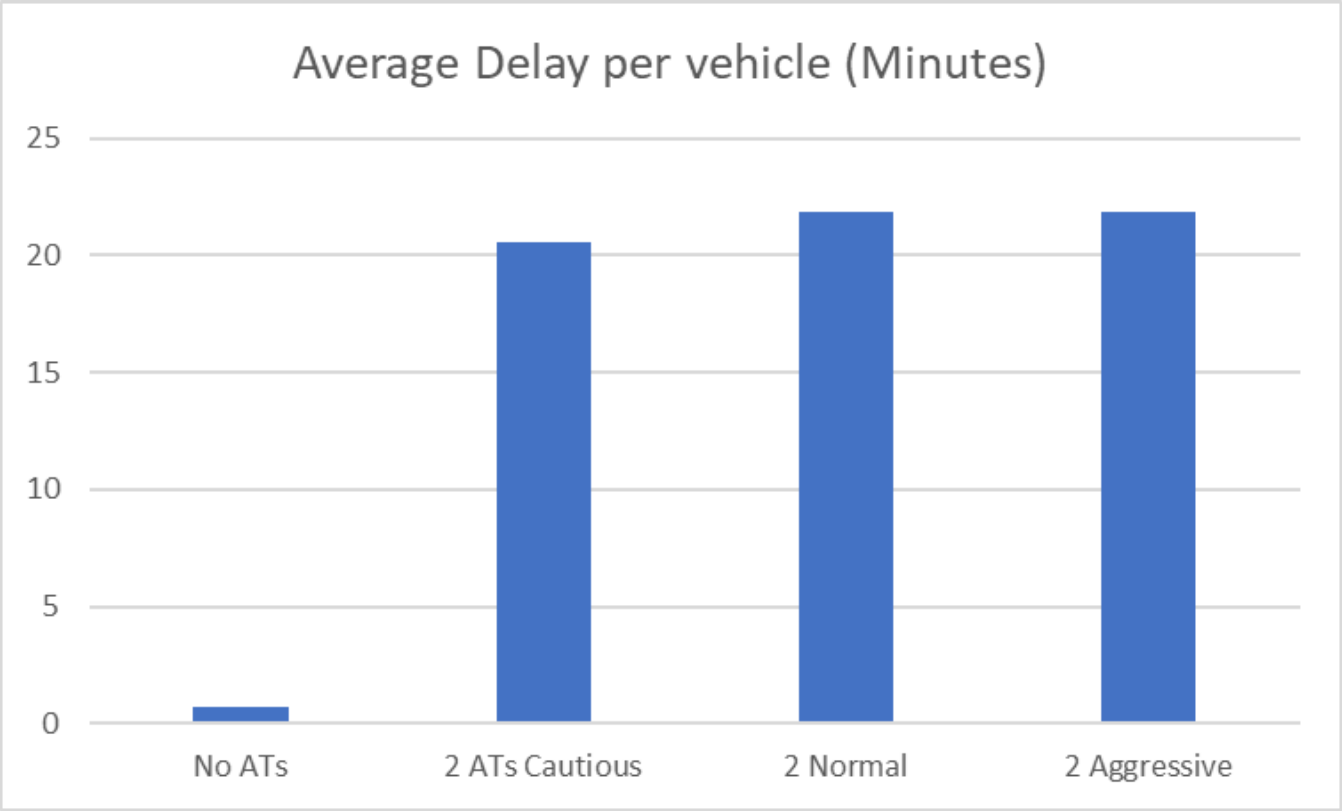

*Figure 7: Average delay per vehicle (Minutes) in Marysville network.*

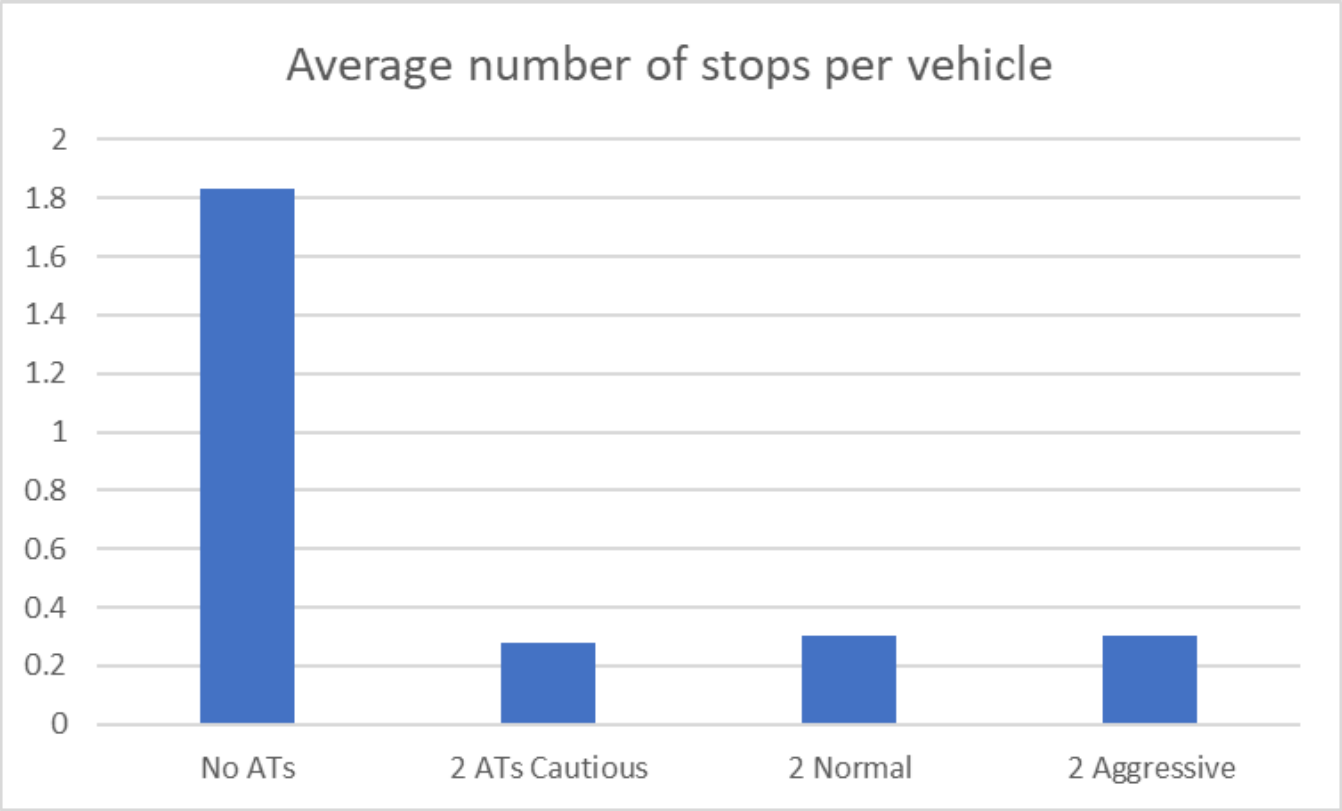

*Figure 8: Average number of stops per vehicle in Marysville network.*

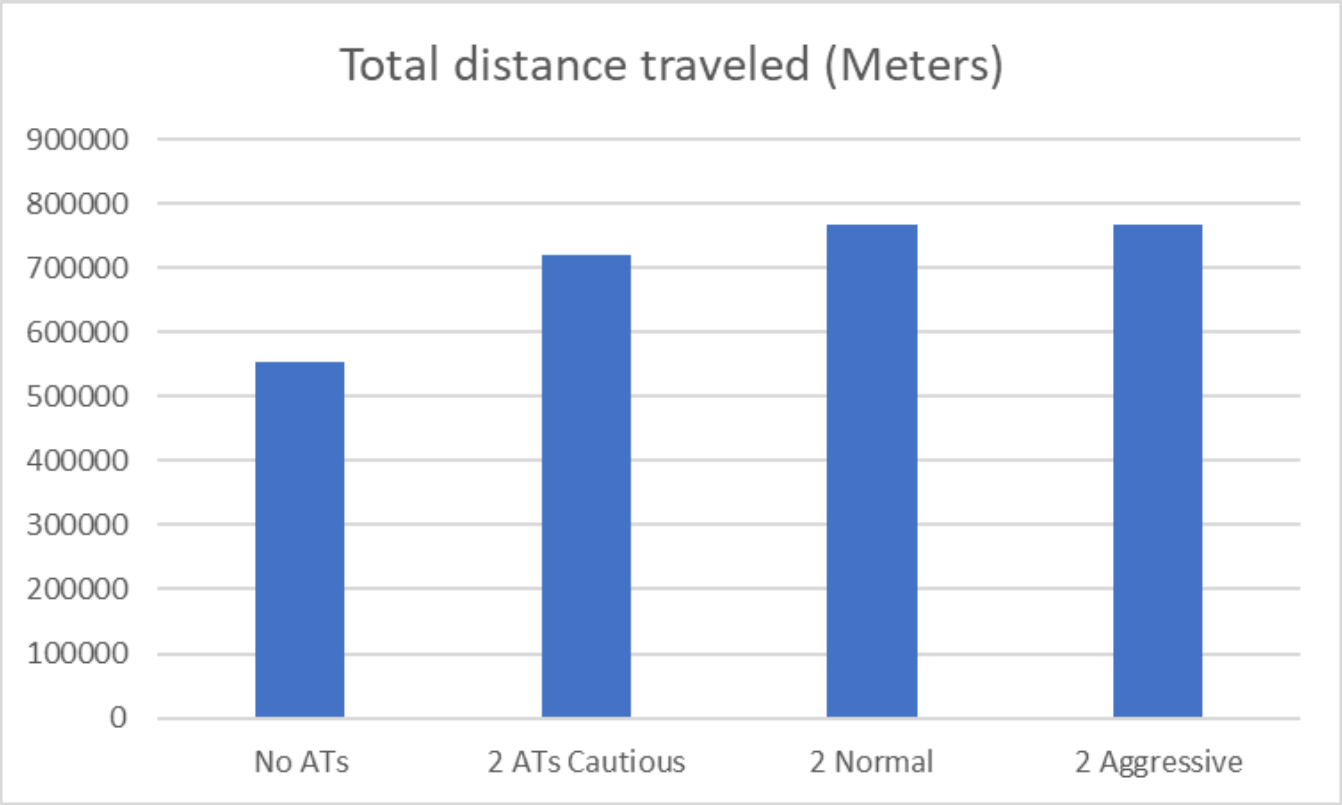

*Figure 9: Total distance traveled (Meters) in Marysville network.*

Further, the simulation results yield average number of completed trips, average trip wait time and total number of passengers served. For these results, the number of completed trips by each fleet of SAV is scaled based on the type of driver and size of fleet. It is noticed that the improvement over a fleet of size 8 and a fleet of size 10 is smaller than that of the improvement between increaing the fleet for 2-8 SAVs. These can be observed in Figure 10.

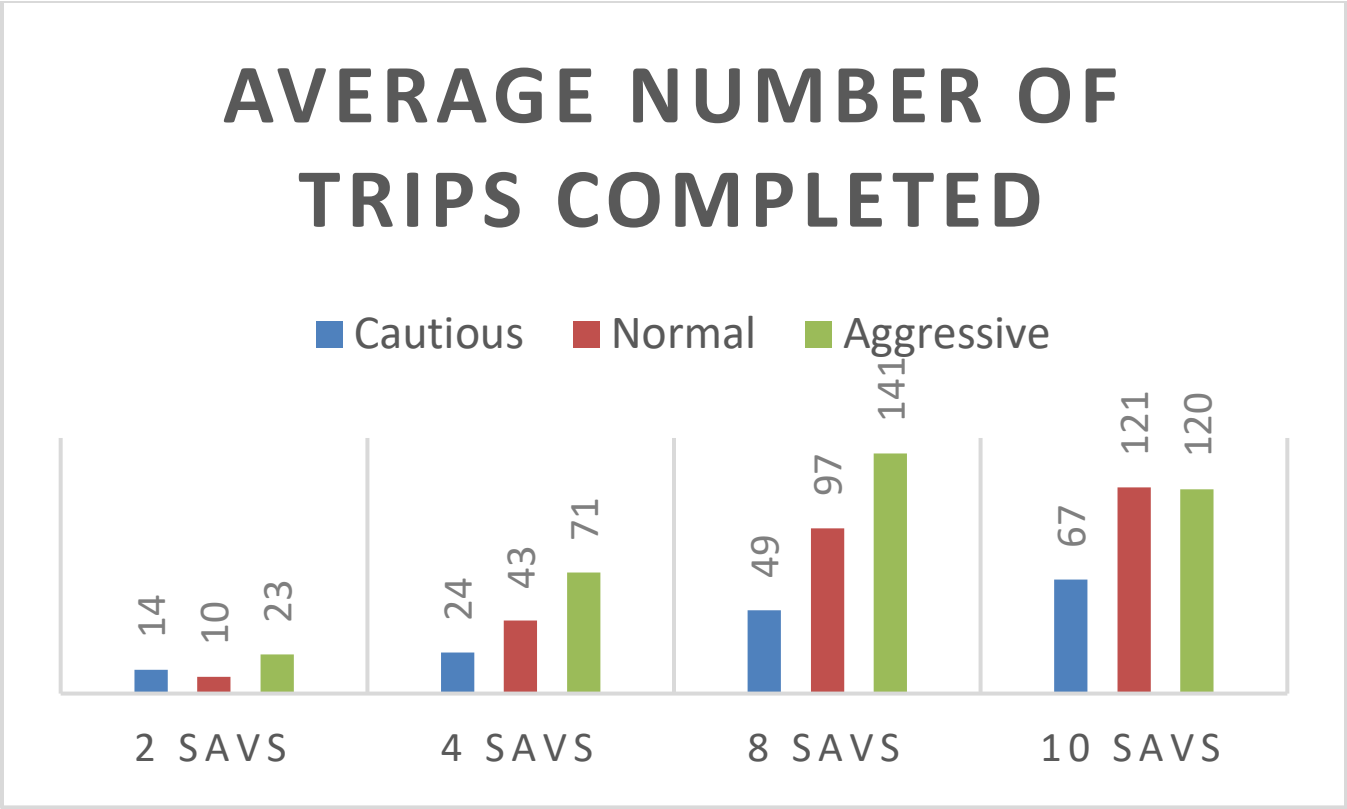

*Figure 10: Average number of trips completed.*

The average wait time for pickup varies from 50 minutes, to in the worst case scenario, 1 hour 40 minutes. This can be attributed to the size of the network. Utilizing the soft deadline of 20 minutes leaves the SAVs with the additional task of driving to pickup passengers. This results in added overhead between commutes. This can be observed in Figure 11.

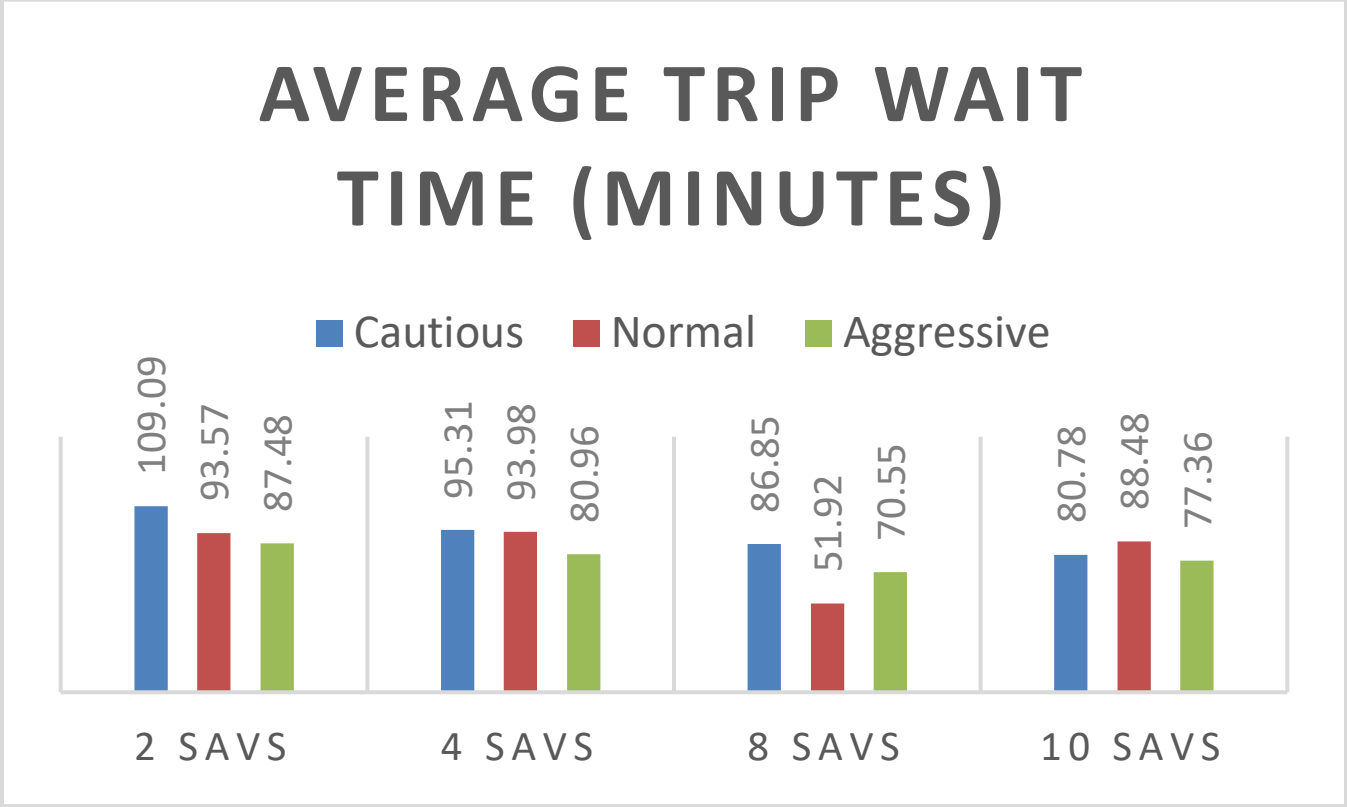

*Figure 11: Average trip wait time (Minutes).*

Finally, the number of passenger served with the fleet is shown to be substantial when the fleet size is 8. Again, the number of passengers served is heavily influenced by the number of trips the SAVs are able to accomplish. In the case of the Marysville network, the size of the network is a bottleneck for the trips. This is shown in Figure 12. The continuation of our work will focus on doing the SAV dispatch using an optimized and learning manner to improve the results.

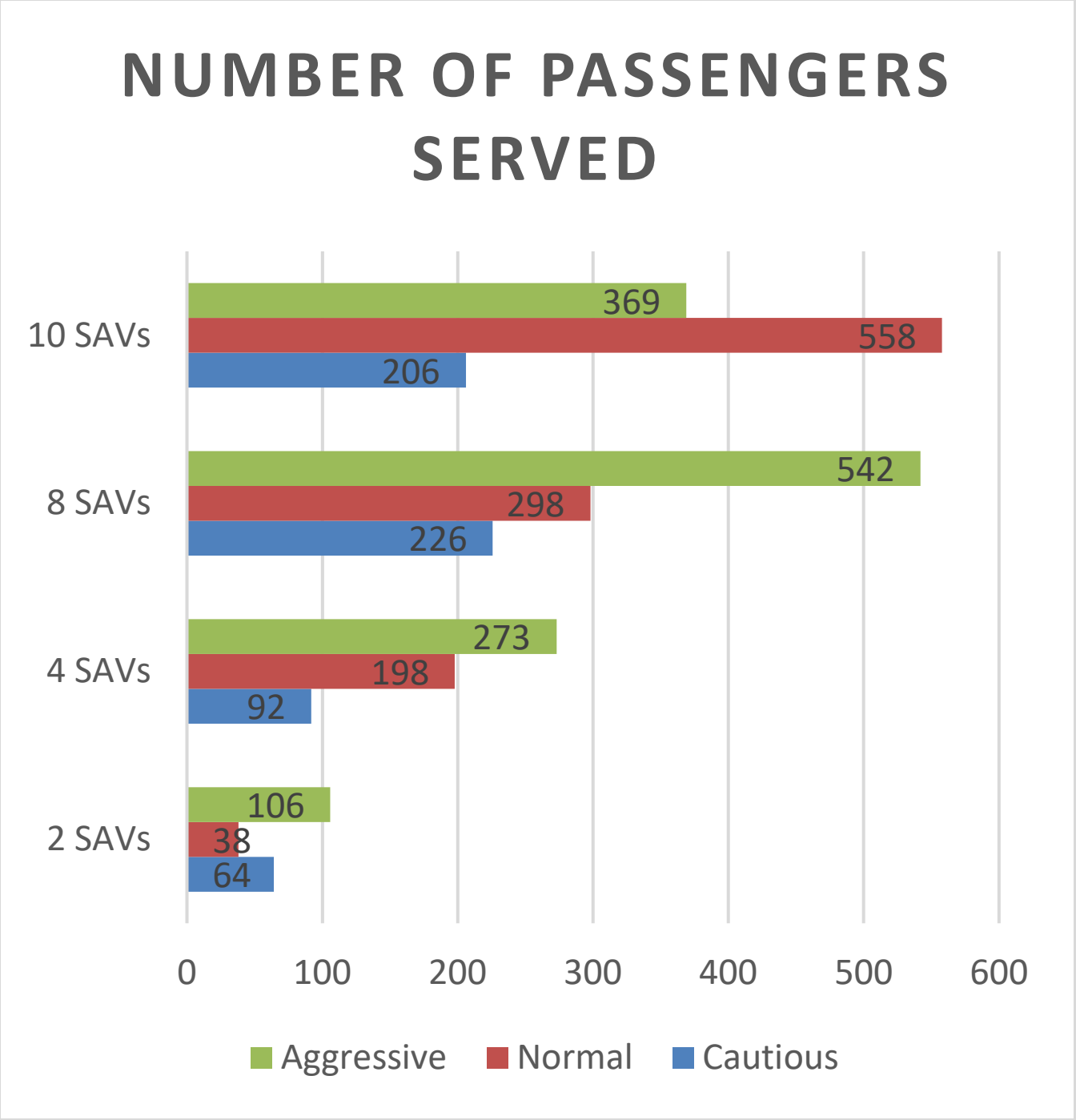

*Figure 12: Number of Passengers Served.*

## Summary/Conclusions

Marysville, Ohio is undergoing a transformation to provide individuals with access to more resources. Part of this transformation lies in enabling its residents to reach destinations to attend school, shop and obtain better jobs. Since there is no public transportion in Marysville and since a large portion of the population do not own or operate private vehicles, the logical solution is to use an SAV mobility service. This makes Marysville a perfect candidate to showcase the effects of the introduction of SAVs in a traffic network with respect to network flow and movement of people. In this paper, the Marysville network was modeled. Utilizing data from the ODOT website, the network was also tuned. Trip requests were modeled based on inference from regular traffic trip times. A baseline of network traffic was introduced, and it was observed that some zones did not receive transport. Further, an SAV service was modeled with high fidelity. Lastly, a couple of ATs were introduced into the network to understand the impact of such, and it was shown that, because of the size of the network, there was no significant overload. In this case, increasing the traffic fleet benefits the network as it serves the areas with no traffic. It was also found that a fleet of SAVs can serve hundreds of people in the network, providing them with access to resources they could not have access to before. Further, it was found that the size of the network was a bottleneck for the performance of the SAV fleet. This can be solved by the addition of further SAV hubs and pickup/dropoff locations. It was also found that the number of SAVs introduced improved the overall performance of the fleet dramatically until the fleet reached size 8. From there on, there was still improvements but they were not as dramatic. Future work includes the refinement of the dispatching service to better attend to requests with the use of reinforcement learning.

## Contact Information

Karina Meneses-Cime

menesescime.1@osu.edu
Automated Driving Lab, Ohio State University
1320 Kinnear Rd., Columbus, OH, 43212

## Acknowledgments

The authors would like to thank Ford Motor Company for partial support of this work. The authors acknowledge the support of this work in part by the National Science Foundation under Grant 2042715.